\documentclass[sigconf, screen]{acmart}
\renewcommand\footnotetextcopyrightpermission[1]{} 
\AtBeginDocument{%
  \providecommand\BibTeX{{%
    \normalfont B\kern-0.5em{\scshape i\kern-0.25em b}\kern-0.8em\TeX}}}

\begin{document}

\title{Automated Isovist Computation for Minecraft}

\author{Jean-Baptiste Hervé}
\affiliation{
  \institution{University of Hertfordshire}
  \streetaddress{College Lane}
  \city{Hatfield}
  \country{UK}}
\email{jbaptiste.herve@gmail.com}

\author{Christoph Salge}
\affiliation{%
  \institution{University of Hertfordshire}
  \streetaddress{College Lane}
  \city{Hatfield}
  \country{UK}}
\email{ChristophSalge@gmail.com}

\renewcommand{\shortauthors}{Hervé and Salge}

\begin{abstract}
Procedural content generation for games is a growing trend in both research and industry, even though there is no consensus of how good content looks, nor how to automatically evaluate it. A number of metrics have been developed in the past, usually focused on the artifact as a whole, and mostly lacking grounding in human experience. In this study we develop a new set of automated metrics, motivated by ideas from architecture, namely isovists and space syntax, which have a track record of capturing human experience of space. These metrics can be computed for a specific game state, from the player's perspective, and take into account their embodiment in the game world. We show how to apply those metrics to the 3d blockworld of Minecraft. We use a dataset of generated settlements from the GDMC Settlement Generation Challenge in Minecraft and establish several rank-based correlations between the isovist properties and the rating human judges gave those settelements. We also produce a range of heat maps that demonstrate the location based applicability of the approach, which allows for development of those metrics as measures for a game experience at a specific time and space.  
\end{abstract}


\begin{teaserfigure}
  \includegraphics[width=\textwidth]{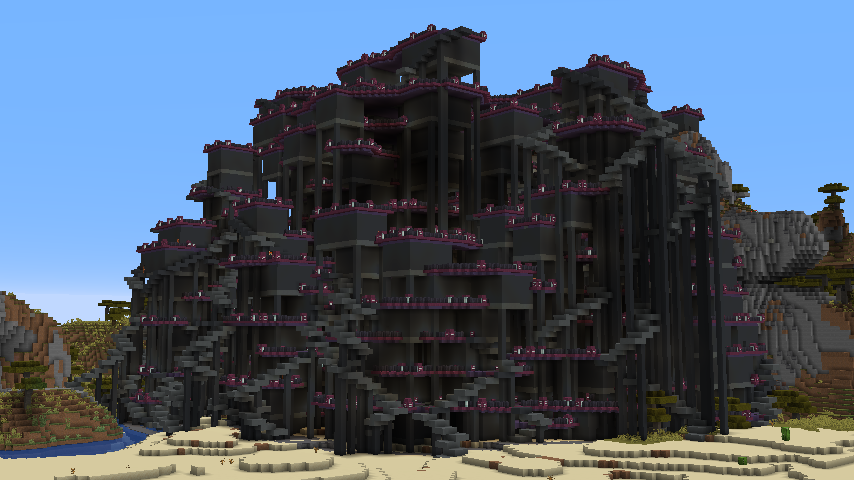}
  \caption{This dystopian city has been generated using Nils Gawlik's submission. It was also ranked as the best looking entry. This generator has been mostly inspired by Kowloon Walled City}
  \label{fig:teaser}
\end{teaserfigure}

\maketitle
\pagestyle{plain}

\section{Introduction}
Can isovist theory from architecture provide us with a metric to automatically evaluate game in general, and procedural generated content (PCG) in particular? Being able to quickly, and without human intervention, evaluate generated content would provide a massive boost to the impact PCG has on game design. It would also methodologically helpful to the academic field of PCG \cite{shaker2016procedural,short2017procedural,katecompton2016,liapis2014computational}, which struggles to provide quantitative data about newly developed techniques. Approaches such as expressive range analysis \cite{smith2010analyzing}, and its extensions \cite{cook2016towards}, provide valuable insights, but focus on measuring the diversity, rather than the quality of various artifacts. A range of existing metrics are suitable to say if two levels or artifacts are similar or not, but struggle to indicate their quality. Some attempts of the past to ground these existing metrics by comparing them to human experience - either of general quality \cite{marino2015empirical,herve2021comparing}, or specific desired experiences \cite{guckelsberger2017predicting,5740836}, are a step in the right direction, but provide mixed results, and showcase several problems with existing metrics - such as some artifacts proving too complex to evaluate on a ``per-level'' basis, or difficulties with transitioning between 2d and 3d.
In this paper, we attempt to develop a new set of metrics, based on theories from architecture, focused on isovists\cite{benedikt1979take}, i.e. the space visible from a given vantage point. This is a quantitative approach to space in architecture, with an established track record to reflect human experiences and behaviour \cite{wiener2007isovist, weitkamp2014validation}. Its agent-focussed definition also allows for metrics that are influenced by the specific embodiment of the agent and are able to provide a metric for an experience of a given moment - allowing for evaluation of not just artefacts as a whole, but also for evaluations of trajectories through the state space of a game, or a specific game state. 
We develop a set-based computation approach, that allows us to apply these measures to 3d discrete environments, such as Minecraft \cite{minecraft} blockworlds - but aim to not incorporate any Minecraft specific features to keep the measures general. As a first study to evaluate these measures, we use a dataset from the GDMC AI Settlement Generation Challenge \cite{salge2018} in Minecraft. We don't focus on the actual generation, and just look at the various published maps and evaluations from human judges. Similar to other PCG AI competitions, such as \cite{khalifa2016general,shaker20112010,stephenson2016procedural}, GDMC uses human judges rather than automatic evaluation, indicating further, that there seems to be no generally agreed upon, automatic, measure of quality.

As human judgement is provided on a ``per generator'' basis, we compare the average score of our measurements with the human judgements, and find some interesting correlations, particularly between the perimeter of the isovist, and the perceived adaptability of the settlement. A measure for visible block types, which we considered as a agent-focussed refinement of the block diversity measure from \cite{herve2021comparing} also correlates well with various human judgements. We also provide location based heatmaps for selected maps and measures, to show how those values would change as the player moves around the map - showcasing the location-based possibilities of the new measures. 
We will now first introduce some more details about PCG and the GDMC challenge, and introduce the isovist concepts and its computation in more detail, before discussing the results and its implications. In general, we conclude that these isovist based measures seem a useful approach to PCG evaluation that warrant further study.

\section{PCG}

PCG is the ensemble of techniques that aim to create game content algorithmically. It has been used to generate content of various nature, from game assets to game play rules. Those techniques are being used in the video game industry, and in the same time constitute a field of research. One recurring challenge in PCG is evaluating the output \cite{smith2010analyzing, summerville2017understanding}. A generator able to self evaluate the content it produces can improve itself or reliably curates which asset is relevant or not, depending of the technique being used. Metrics of evaluation can also be used by designers to tune the generators and optimize certain aspect of the generated artifacts. 

The core constraint is usually playability - aimed to ensure that a game can actually be used as such. But other metrics have been developed with the intent to evaluate other dimensions of an artifact, such as its looks, the narrative it conveys, its impact on game mechanics, or even how "fun" it is. However these metrics tends to lack of human grounding \cite{marino2015empirical, herve2021comparing}. They are also build to focus the entirety of the artifact, and are rarely designed for a local use, targeting for instance a whole level instead of a single location.

Therefore, developing new metrics and improving existing ones is necessary in order to improve PCG as a field, for various reason (automatic curation, co-authoring, ...). Beyond the question of the metrics themselves lays another one : How to properly use them? In the current paper, we try to address both of these concern, by establishing a new range of metrics, polishing an existing one, and compare their efficiency both globally and locally.

\section{GDMC}

 Minecraft \cite{minecraft} is a voxel based game developed by Mojang Studio, where the players progress in an open world made out of blocks. These blocks represent different materials, such as wood, rock and so on. Players can destroy blocks, place them in any position within the world, or even combine them through crafting mechanics in order to create new types of block or item. Minecraft is mostly known for its open-endedness and is mostly used as a sandbox game. Many players use the blocks mechanics to terraform the game world, create structures such as houses, castles or cities. Since the art style and the setting of Minecraft is very generic, the game affords free creation of almost any kind of artifact, with only the player's imagination setting the limits.
\par
The Generative Design in Minecraft Competition (GDMC) is a PCG competition in which competitors submit a settlement generator \cite{salge2018}. All the submitted generators are then tested on fixed maps, which are selected by the organizers \cite{salge2020ai}. All the generated settlements are then sent to the jury. This jury includes experts in various field, such as AI, Game Design or Urbanism. Each judge scores in each of the following categories : \emph{Adaptability, Functionality, Narrative, Aesthetic}. \emph{Adaptability} is how well the settlement is adapted to its location - how well it adapts to the terrain, both on a large and small scale. \emph{Functionality} is about what affordances the settlement provides, both to the Minecraft player and the simulated villagers. It covers various aspects, such as food, production, navigability, security, etc. \emph{Narrative} reflects how well the settlement \emph{itself} tells an evocative story about its own history, and about who its inhabitants are (There is a separate bonus challenge about also adding a written PCG text that tells the story of the settlement\cite{salge2019generative}). \emph{Aesthetic} is a rating of the overall look of the settlements. In the competition, the rating of each category is computed for each generator by averaging (mean) across all judge's scores. The judges provide for each generator, after looking at every maps, one score for each of the four categories.
The overall score of the generators is then obtained by a mean average over the four categories.

The human data we are working with are the average scores for the generator. We therefore have, for each generator, 5 scores: the overall score, adaptivity, functionality, narrative and aesthetics.
In 2021, the competition received 20 submissions.

\section{Theory of space}
\subsection{Isovist}

Given a bounded environment, for each point $x$, we can compute its isovist $V_x$ \cite{benedikt1979take}, which is the set of all the points visible from $x$. $x$ is called the centroid of the isovist. The lines connecting $x$ and the boundary of $V_x$ are referred to as radials $l_{x,\theta}$.
Based on $V_x$, we can compute several properties characterizing it.
First of all, the visible area ($A_x$), and the perimeter ($P_x$). It is worthwhile to note that $P_x$ is defined by "real-surface" which are defined by Benedidkt as “opaque, material, visible surface, humanly perceivable as such" \cite{benedikt1979take}, and therefore exclude the sky or any glass surface from the computation. In addition to $P_x$, the occlusivity ($Q_x$) contains not only real-surfaces, but also any vision blocking surface which are not perceived as such. The remaining properties are focused on the radials, with their Variance, Skewness (asymmetry of the radials) and Circularity of $V_x$.

\begin{figure}[htpb]
\centering
\includegraphics[width=2.5in]{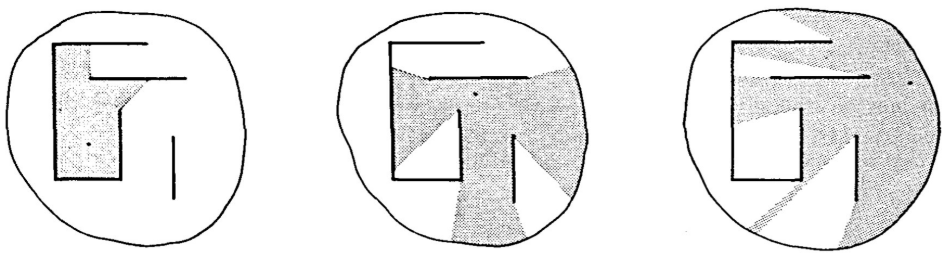}
\caption{Example of a set of 3 different isovist in a single environment. From \cite{bresenham1977linear}}
\label{fig_Isovist}
\end{figure}

\subsection{Isovist and Experience of Space}

Isovists are particularly interesting, since they have been developed with the purpose of capturing perception of space \cite{benedikt1979take}, and have been proved to have some correlation of how a given space is experienced and appreciated \cite{wiener2007isovist, weitkamp2014validation}.

Space in general can be analyzed to understand and even predict human and social behavior. The Space Syntax theory \cite{hillier1976space} has been used to study the features of any kind of human environment (inside of a building, mall, city,...), how it has evolved and how it is used \cite{ortega2005space, van2017space}. Space syntax analysis integrates isovist, and refer to it as “isovist space” \cite{turner1999making, turner2001isovists}. While a space syntax analysis of Minecraft’s settlement is out of this paper’s scope, it will definitely be an interesting follow up.

We are particularly motivated in investigating space as a key component of any 3D game\cite{nitsche2008video}. While it is obvious that space characterizes the gameplay experience \cite{nitsche2008video}, it is also deeply connected with the narrative \cite{fernandez2011spaces, nitsche2008video} and social experience \cite{nitsche2008video}. Our hope is that understanding space is a step in understanding all the layers that compose a video game.
\subsection{3D Isovist}

In its definition, isovist can be used in 2D environments as well as in 3D, and several publication focus specifically on that topic alone \cite{ giseop2019new, suleiman2013new}. However, we found very few practical attempts to use these method in a 3D environment \cite{psarra2015experiencing}. Using the isovist as defined in the first place was an option, but given that Minecraft is a special environment, with its own vision and movement rules, we redefined some of them.

First of all the Perimeter. Vision in a virtual environment is not necessarily the same for each and every player. Depending on your hardware, the view distance, screen resolution and field of view angle (FOV) will vary. Three Minecraft players can have a different view of distant mountains: one might sees nothing, the second its silhouette, while the third one may scrutinizes it in full details. Because a Minecraft environment has no clear boundaries, we decided to include in our perimeter every end of our radials. This grants a larger perimeter to outside isovists in open fields. It is an attempt to compensate for the fixed view distance we use in our computations.
It is also worthwhile to note that while the Perimeter value usually works perfectly to capture features such as walls, in Minecraft we do not have any tool to characterize the purpose of a vision blocking element. A  block is a block, and its name or position cannot determine if they are part of a floor, a wall, or a ceiling.
A good example would be roofs. If you were to look at a house built in Minecraft, you will probably be able to characterize which blocks are part of a house. But from a computational point of view, this is a much more trickier task. You can technically walk on a roof as if it was part of the environment. You cannot rely on the type of block used, as anything could be technically used, neither the position of the block relative to your isovist, since it could be the top of a tree, a hill, or any other environmental feature. It would be possible to specialize our algorithm for Minecraft game’s mechanics, but the result would likely stays imprecise while making the evaluation technique less generic, which goes against the aim this paper.
For all these reasons, our Perimeter is composed of any vision blocking blocks. We however excluded doors from this list, as they might be opened, letting players see through.
We also compute another metric, called Real Perimeter, which is the subset of opaque blocks in our perimeter (mostly removing the air blocks at the maximum vision range of our radials). In other isovist terms, we are only counting the real surfaces with this metric, making it closer to it original definition.

\begin{figure}
    \centering
    \includegraphics[width=7cm]{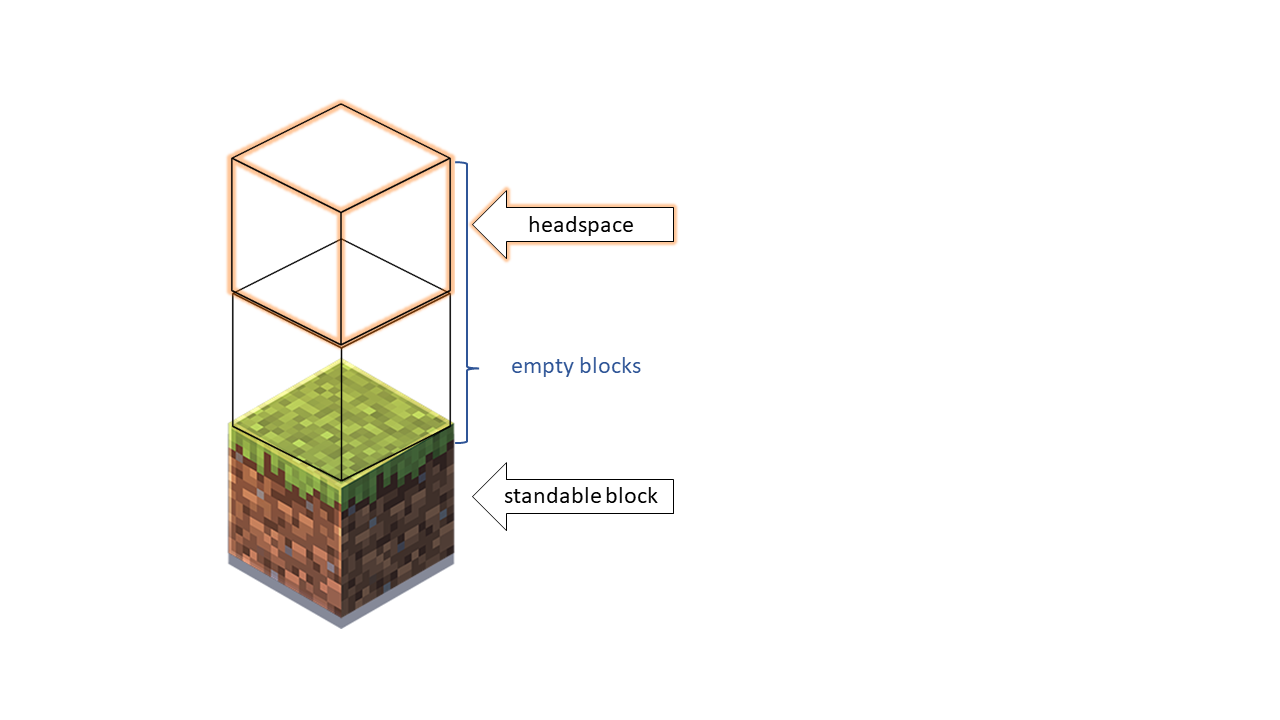}
    \caption{Illustration of the headspace concept.}
    \label{fig:headspace}
\end{figure}

The Area is another property that we tuned. Isovit’s Area could be defined as: every point visible from the centroid, and from which the centroid is visible. This reciprocity works well in 2D and on flat surfaces, but works differently in Minecraft’s noisy worlds. In numerous scenarios, you may not be able to see a close by surface where your in game character could stand. However, knowledge of the game’s mechanics lets you guess the potential “surfaces” in your surroundings. It could be the shape of the terrain, but also looking at another player’s head in a configuration where you cannot see their feet. We therefore defined the Area as all the headspaces visible (a headspace being a spot where a player’s head would fit if they were to stand on a surface 2 blocks below, see Fig.\ref{fig:headspace}).

We measure both Variance and Average of the radials, but discarded the skewness, since its relevance was not underlined by the literature we explored. We also keep track of the longest radial, which is often refereed as the Vista Length. In place of the Circularity, we approximate two values: Roundness, the ratio between the Area and the Perimeter, and Openness being the ratio between Area and the Real Perimeter. Openness is another measure, frequently used along isovist \cite{weitkamp2014validation, psarra2015experiencing, giseop2019new}, and defined as "the amount of space visible by the viewer" \cite{kaplan1989environmental}. While isovist focus on the surface and features blocking the vision, the Openness is targeting the volume of visible space, regardless of it significance (e.g. Isovist does not include the sky in its computation).

Another metric added to the isovist theory later on is the Drift \cite{dalton2001omnivista}. An isovist's drift can be defined as the distance between the centroid and the 'center of gravity' of its Area. An isovist close to a wall would end up with a high drift value, while an isovist in the center of a squared room will tend to have a lower drift. Computing the exact center of gravity in our case presented several challenges, but we approximated it by averaging all the vectors of our radials end point.

We developed another metric in an attempt to tie what the player sees and the game locomotion, which is the Reachability. We defined it as the part of the Area accessible to the player by simply walking, with no other game specific mechanics involved. In Minecraft specifically, it corresponds to the section of the Area reachable without destroying or placing blocks. Reachability is computed using a floodfill algorithm, starting from the isovist's centroid, with a fixed amount of steps.

Occlusivity makes less sense in the current context, since all the vision blocking blocks can be considered as real surface. But we tried to adapt it based on the Reachability, and how much of that accessible space is visible. Following on this idea of visible spot were a player can stand, we also introduce the Clutter metric, and define it as the ratio between the reachable headspaces entirely visible and the Area.

Finally, since previous Minecraft focused research has highlighted the variety of block types as a relatively good metric \cite{herve2021comparing}, we also compute the number of visible block types for each isovist.

\subsection{Set-based definitions of isovist properties}

In this section we outline a set-based way to define and compute the various, above mentioned properties for a Minecraft-like 3d blockworld. This has several advantages. It should give a clear, and reproducible definition. It should provide an instructive account of how to efficiently implement the values, reusing a limited amount of pre-computed sets for each location for all values. Finally, it will provide conceptual clarity of were various forms of embodiment and game play parameters enter the computation. 

A Minecraft world can be, in large part, defined by its constituent blocks. Each 3d coordinate basically contains one block, and defining the type of block present at each coordinate defines the world map. Other objects are also made up of blocks, just arranged differently. 

We will define a number of sets, denoted by capital letters $X$, each containing a number of unique blocks defined by their x,y,z coordinates and their type. 

We use the concept of \emph{headspace}, defined as the block a player avatar's head could be for a standing avatar. Every block that is empty, has an empty block below it, and has a \emph{standable} block below that, is a possible headspace. The list of empty blocks contains air, but also some less common blocks such as doors, carpet, etc, that still allow the player to enter those blocks. Standable blocks are similarly defined via a list, and contain most solid blocks, but exclude things like lava or water. The two empty blocks are due to the 2 unit height of a player avatar. This should capture all blocks a player could be in $H_{\forall}$. The following sets are computed for all possible headspaces $x \in H_{\forall}$.
\begin{eqnarray}
   H_{x,d} &:=& \{\textrm{all } x \in H_{\forall} \textrm{ visible from } x\}  \\
   H_{x-2} &:=& \{\textrm{blocks 2 units below every } \in H_x\}  \\
   P_{x,d} &:=&\{ \textrm{all blocks visible from } x\}  \\
   Pr_{x,d} &:=&\{ \textrm{all real surfaces blocks visible from } x\}  \\
   R_{x,n} &:=&\{ \textrm{walkable blocks from $x-2$ in $n$ steps}\}
\end{eqnarray}
Visibility between two blocks is computed by ray-casting via Bresenham algorithm \cite{bresenham1977linear}, and checking if the blocks along the ray are transparent, i.e. in the transparency list. Air is the main transparency provider, but notably glass provides transparency without allowing avatars to pass. The parameter $d$ is used to limit the max length for this ray cast. We also compute the length of the many rays cast during the visibility computations to compute:
\begin{eqnarray}
\textrm{var Radials} &:=& Var(l_{x,\theta})\\
\textrm{mean Radials} &:=& Mean(l_{x,\theta})\\
\textrm{Drift} &:=& l_{x,mean(\theta)})\\
\textrm{Vista Length} &:=& Max(l_{x,\theta})\\
\end{eqnarray}
The visible head spaces $H_{x,d}$ are basically all positions an avatar could be in and then see its head from its current position. $H_x-2$ are all the standable blocks supporting those head spaces. $P_{x,d}$ provides a perimeter of blocks that limit our view, and contains all blocks visible from the current position. $R_{x,n}$ is a list of all walkable block, obtained by floodfilling from the standable block supporting the current position, within $n$ steps. We use usual Minecraft movement rules, that allow moving up by one block per lateral transverse, and dropping down to lower levels. Note how the features of avatar height, movement rules and vision sensors could affect those basic sets. The following properties can now be computed by operating on those set alone, without having to recompute them. 

\begin{eqnarray}
\textrm{Area} &:=& |H_{x,d}| \\
\textrm{Perimeter} &:=& |P_{x,d}|\\
\textrm{Diversity} &:=& c(P_{x,d})\\
\textrm{Real Perimeter} &:=& |(Pr_{x,d})|\\
\textrm{Roundness} &:=& Area / Perimeter\\
\textrm{Openness} &:=& Area / Real Perimeter\\
\textrm{Reachability} &:=& |R_{x,n}|\\
\textrm{Occlusivity} &:=& |R_{x,n} \cap H_{x-2}|/ Reachability\\
\textrm{Clutter} &:=& |H_{x-2} \cap P_{x,d}|/ Area\\
\end{eqnarray}
The function $c(.)$ counts how many different types of blocks are in a set. 
\section{Experiment}
\begin{figure*}[htpb]
\centering
\includegraphics[width=\linewidth]{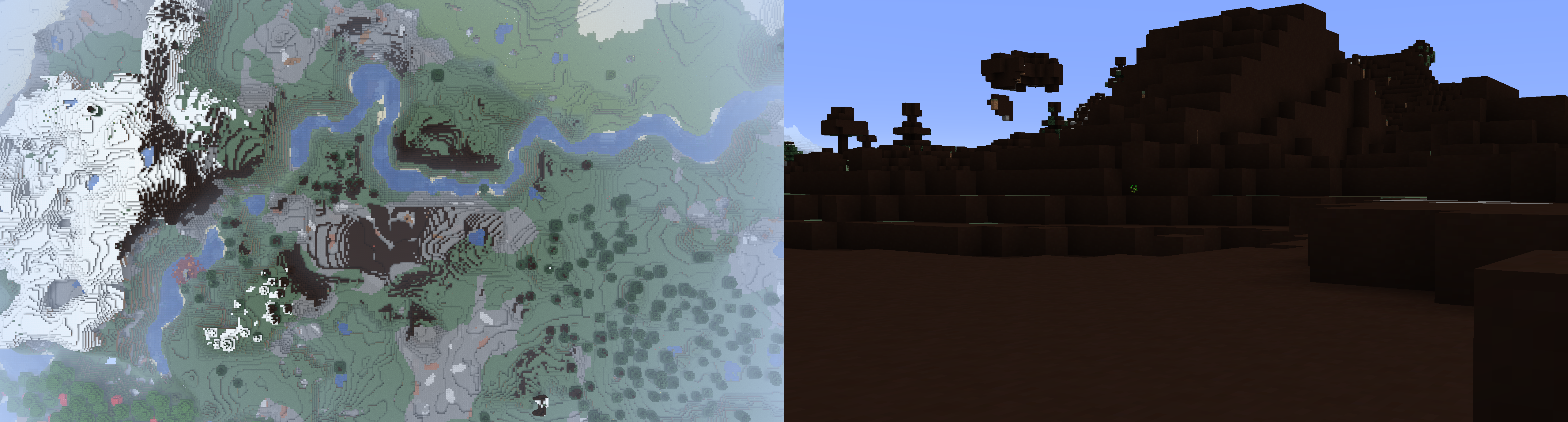}
\caption{Screenshot of an isovist in Minecraft, with perimeter blocks replaced with black blocks. The centroid is in the center of the left picture (top view). The right picture is the vision from the centroid - note that nearly all visible blocks are black, indicating that the bresenham ray cast hits nearly everything visible to us.}
\label{fig_MinecraftIsovist}
\end{figure*}
Our experiment runs on the 20 entries from the 2021 GDMC competition. The 2021 competition was held on 2 maps, and we used both of them.
We compute isovists on surfaces where a player can stand, at a height matching the camera position for the given location. For each isovist, we generate the coordinates of a sphere with a radius $d$ of 256 blocks (the by default distance view in Minecraft), with the same centroid as the isovist. The sphere itself is computed as if it was made out of blocks, for a total of 682746 blocks. We then compute the radial between every sphere’s blocks and the centroid, using a 3D Bresenham algorithm \cite{bresenham1977linear}, and for each coordinate we check the type of block we are currently crossing, if it is \emph{transparent}, and if it is a \emph{headspace} (see Fig.~\ref{fig:headspace}). As an output, we get our Perimeter and our Area, and can derive the rest of the metrics. Our Reachability values are computed using a 10 steps floodfill algorithm.

Despite multiple optimizations, the code still has a significant computation time. The default map, without any structure, has 92560 possible surfaces, and a single isovist takes several seconds to be computed, depending on its size. We therefore proceeded to the following subsampling of our data: for each height value, we compute 1 isovist out of 10 possibles, picked randomly.

Once we gathered the isovist values, we computed the mean average values for each settlements, combining the values from both map. Based on those results, we calculated the ranked based Spearman correlation between the average of isovistic properties and the rating the settlement got in every categories (See Table\ref{tab_correlations}) \footnote{\url{https://docs.google.com/spreadsheets/d/1ZtPX-nUfbx_f6HOYXWvReNlPuLpRvUbo}}.
\section{Results and Discussion}
\subsection{Correlations}
\begin{table*}[htpb]
\caption{Rank Based Correlation between judging criteria and isovist properties.}
\begin{center}
\begin{tabular}{|c||c|c|c|c|c|}
\hline
&Adaptability & Functionality & Narrative & Aesthetic & Overall\\
\hline
\hline
Area &  0.28 & -0.22 & -0.26 & -0.31 & -0.17\\
\hline
Perimeter & 0.59 &	0.12 & -0.16 &	-0.25 & 0.09\\
\hline
Diversity & 0.21 & 0.6 & 0.47 & 0.66 & 0.61\\
\hline
Var Radials & -0.48 & -0.16 & -0.01 & 0.28 & -0.14\\
\hline
Mean Radials & 0.004 & -0.03 & -0.33 & 0.02 & -0.11\\
\hline
Roundness & -0.33 & -0.48 & -0.1 & -0.27 & -0.39\\
\hline
Openness & 0.59 & 0.04 & -0.22 & -0.35 & 0.02\\
\hline
Clutter & -0.14 & 0.16 & 0.31 & 0.23 & 0.19\\
\hline
Reachability(10) & -0.40 & -0.43 & -0.08 & -0.21 & -0.37\\
\hline
Occlusivity & 0.46 & 0.06 & -0.19 & -0.25 & 0.03\\
\hline
Drift Length & 0.18 & 0.25 & -0.07 & 0.18 & 0.18\\
\hline
Vista Length & -0.11 & -0.06 & -0.22 & 0.1 & -0.1\\
\hline
Real Perimeter Size & 0.05 & -0.13 & -0.15 & -0.1 & -0.11\\
\hline
\end{tabular}
\label{tab_correlations}
\end{center}
\end{table*}

\begin{table*}[htpb]
\caption{p-value between judging criteria and isovist properties.}
\begin{center}
\begin{tabular}{|c||c|c|c|c|c|}
\hline
&Adaptability & Functionality & Narrative & Aesthetic & Overall\\
\hline
\hline
Area & 0.22 & 0.36 & 0.28 & 0.19 & 0.47 \\
\hline
Perimeter & 0.01 & 0.63 & 0.49 & 0.28 & 0.72\\
\hline
Diversity & 0.38 & 0.005 & 0.04 & 0.002 & 0.004\\
\hline
Var Radials & 0.03 & 0.51 & 0.97 & 0.23 & 0.56\\
\hline
Mean Radial & 0.99 & 0.9 & 0.16 & 0.93 & 0.65\\
\hline
Roundness & 0.16 & 0.03 & 0.69 & 0.26 & 0.09\\
\hline
Openness & 0.01 & 0.87 & 0.34 & 0.13 & 0.92\\
\hline
Clutter & 0.56 & 0.5 & 0.19 & 0.32 & 0.42\\
\hline
Reachability(10) & 0.08 & 0.06 & 0.72 & 0.37 & 0.11\\
\hline
Occlusivity & 0.04 & 0.79 & 0.43 & 0.29 & 0.9\\
\hline
Drift Length & 0.44 & 0.3 & 0.77 & 0.46 & 0.46\\
\hline
Vista Length & 0.64 & 0.82 & 0.36 & 0.68 & 0.68\\
\hline
Real Perimeter Size & 0.84 & 0.58 & 0.53 & 0.69 & 0.65\\
\hline
\end{tabular}
\label{tab_pvalues}
\end{center}
\end{table*}
What we observed at first is that the visible Diversity of block types, which is derived from an existing metrics \cite{herve2021comparing}, is having the best performance. We establish several significant (with p-value $p<0.01$, see Table.\ref{tab_correlations} and Table.\ref{tab_pvalues}) relations between the Diversity and the following rating: Functionality, Aesthetic, and the Overall score. To a certain extent ($p=0.04$), it also slightly correlates with the Narrative component of the settlement.
The Perimeter and Openness correlate with with the Adaptability (both wih a $p<0.01$). 
There are a range of further medium significant correlations ($p<0.05$), but we note that given the range of correlations we computed, those run a risk of being spurious. Namely, Radials Variance and Occlusivity seem to have a connection with Adaptability. Finally, the Roundness($p=0.03$) and Reachability ($p=0.055$) seem to correlate slightly with the Functionality of the settlement.

The Diversity of visible block types metrics is an agent-centric reformulation of block type count \cite{herve2021comparing}. The previous metric counted the overall number of different blocks in a settlement, and had the best correlations with human quality judgments. Seeing this affirmed from an agent's perspective is expected. In general, it makes sense that a settlement that produces a range of different affordances\cite{cardona2013cognitivist} uses, and displays, a number of blocks. Noteworthy is that the correlation to the Narrative score diminished compared to the previous work. A possible explanation is the more widespread use of furnishing for the 2021 generators. In previous years furnishing was reported factor for narrative scoring \cite{salge2021impressions}, and furnishing usually introduces the use of different blocks than building. Widespread furnishing likely made this effect less pronounced. Regarding the Aesthetic and Overall categories, it is possible that this metric actually captures those two dimensions, but it is also possible that well polished generators tend to produce a larger variety of blocks. In opposition, a generator that struggles with building a simple house might not have been designed with the ability to use a wide range of assets. Also, the same objection outlined before is still relevant here - this seems to be a measurement highly indicative of perceived quality, but not one that could be maximized computationally in order to achieve an ideal human-experience result by itself.

The relation between Adaptability and Perimeter, and Openness, is encouraging, as none of the previously tested configuration-based metrics in \cite{herve2021comparing} showed any significant correlations with human judged quality. Similarly, \cite{marino2015empirical} also only found count based metrics with good grounding.
In our definition, the Perimeter is increased by the absence of walls and roofs, and therefore and outdoor isovist will have a large Perimeter. This phenomenon is highlighted in Fig.~\ref{fig_heatmaps}, where the buildings with roofs are easy to spot. Another noticeable aspect of this metric is that the bare maps have an average Perimeter higher that any of the entry. It seems that the Perimeter does not capture how well integrated in the environment the structures are, or how much the terrain has been modified, but rather indicates the presence or not of artificial structures, which produce vision blocking surfaces.
We observed that the Openness value works as following: low Openness area are usually in corridors like structure, where most of the vision is blocked by real-surfaces (walls and roofs), while offering very few Headspaces. On the contrary, high Openness isovist are close to a wall-like structure, blocking part of their vision and therefore diminishing the amount of visible real surfaces, but also having a view on large open area. Even if this behavior does not capture the whole concept of Openness as it is commonly used, it successfully highlight layout (e.g. corridors) occurring rarely in natural environment, and increased by artificial structures. As for Perimeters, Openness captures successfully the presence of artificial structures, which is part or the Adaptability of a settlement, but it as no consideration of how well these structures are integrated in their environment.
The other potential correlations, with Radials Variance and Occlusivity, are also motivating us to fine tune these metrics in further work. 

\subsection{Heatmaps}

\begin{figure*}[htpb]
\centering
\includegraphics[width=\linewidth]{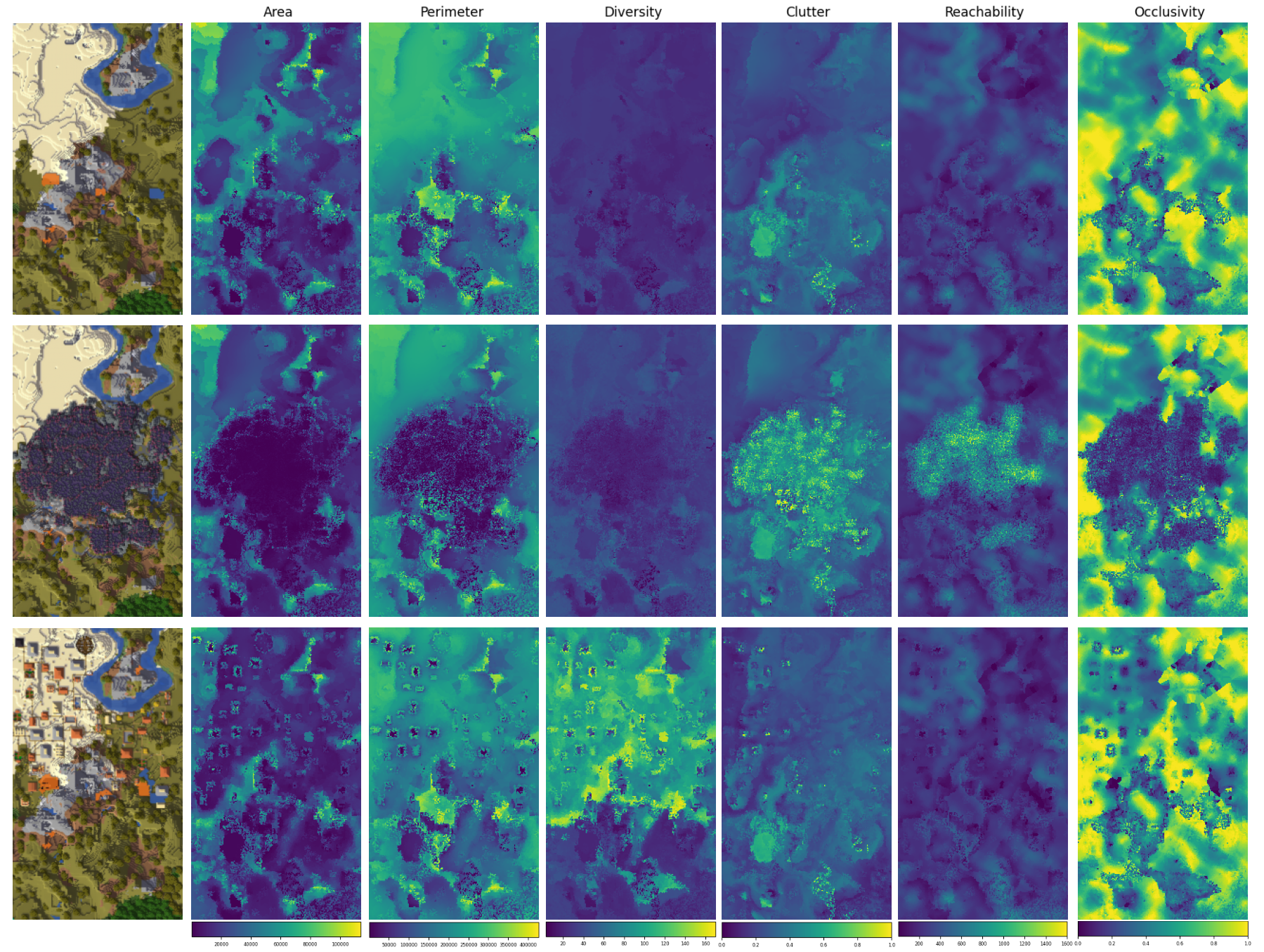}
\caption{From top to bottoms, top-down heatmaps of the base map (volcano type), best Aesthetic scoring map (dystopian city block) ans winning map (typical fantasy settlement). Only the isovists above ground (y$>$60) are represented, with x,z coordinates without measured values being interpolated by nearest neighbors.}
\label{fig_heatmaps}
\end{figure*}

Given the locality of our isovist values, instead of only looking at the average, we can look at those values for specific coordinates. Fig.~\ref{fig_heatmaps} shows a top-down heatmap for various values that averages all values of a given metric for specific $x,z$ coordinates. We exclude (only for the visualization) values with $y<60$, as underground isovists in natural caves are constant, and confuse the image. We interpolate to the nearest value for those coordinates that have no values due to subsampling. For visual precision sake, we also recomputed the data with another random subsampling rate of 1 out of 2 isovist. First, lets look at the values for Area, Perimeter and Diversity in Fig.~\ref{fig_heatmaps} for both the base map, and the judged best looking map, and the judged best map. One of them features a more typical spread out fantasy village, while the other builds a massive `cyberpunk' city block with a complex layout - see Fig.~\ref{fig:teaser}. 

First, by just looking at the basemap, we can see that our metrics produce indeed local values, which capture salient details about the environment. Area being a good proxy for the height-map, with the volcano in the middle, and the river being nicely visible. We also see the impact the different biomes have on the values, with the difference between dessert in the north and the Forest in the south leading visible in a higher block type count, due to more diversity in the Forest, and a much more smooth perimeter in the featureless dessert in the north. Similarly, the impact that the generated settlements have is also quite evident, allowing for a quick identification of where the settlement is, and its density. 

If we are looking at Diversity we can see that the diverse and widely visible structure by the winning settlement increase the local Diversity values in comparison to the base basemap, offering more interesting visuals across most of the map. This is relevant to note - as when we discussed how we should average the measured values we considered only using the values from inside the build up area, or only from specific areas of the map, so the unchanged wilderness or underground area does not dominate the values. While the exact amount of this difference might vary based on generators, it seems likely that the amount of artificial blocks will be more numerous than the natural ones, thereby providing a proxy for visible human footprint. But we decided to use all isovists for the average, a.) because on average any local effect should still be visible, and b.) there might be an effect on the non-modified area. This was to capture our intuition that the experience outside the settlement might still be affected by a settlement in proximity. For example, a tranquil lake might be less enjoyable with a factory blocking your view of nearby mountains. Also note that the metric for Diversity is now location specific, so we can actually see that there are more things to see close to the village in the north then in the southern, mostly untouched forest, and it underline the spatial boundaries of the experience offered by the settlement. 
Diversity and Aesthetic have the strongest correlation we found during our experiment, but the heatmap of the best looking map is quite similar to the base map, it suggest that while this metric is quite solid in most of the case, there is more about this specific map. But again, the surrounding of the settlements are highlighted, proving the visual impact of the settlement on the map. It also is a good example of a non-local metric, i.e. one were building something half a map away will still influence the value. In contrast, other values such as occlusivity, or clutter, only get influenced by changes in the environment that are very close to the measuring agent.

We can also gain insights into specific settlements with these values. Consider the third placed map, with its massive, dystopian housing block atop the volcano. We can see that the inside of the building massively lowers the seen Area, Perimeter and even Diversity, while those values are in fact elevated at the outer edge. Here the local value might offer insights on where in this structure you would want your apartment to be - and align with peoples usual preferences.

The Clutter metrics also offers interesting heatmaps. This metric has been developed as a proxy for the environment readability. Even if it does not correlate with any judging criteria, it seems to be capturing local features nonetheless. The flat top of the volcano, overlooking the rest of the map, has an higher Clutter ratio than the desert's dune. Regarding the settlements, the large open floors of the megalopolis tower over not only the environment, but the city itself, with the higher floors having the best performing ratio. The city layout allows the players to project themselves in most of the location below them. On the contrary, the more traditional village offers less uniform values. After checking in game, one explanation could be that opening such as windows, give view on only a portion of the outside, with the inability to see part of the ground. This behavior matches the implementation of the metric, and point out one of its limitations: the players do not need to see the entirety of space to understand its layout and properties. But it does underline areas of uncertainty nevertheless, which could be a potential use case.

The measure of ``Reachability'' for a given location is basically the discrete, 10-step empowerment \cite{jung2011empowerment}, as measure usually used to capture how much influence an agent has over the world it perceives. Its noteworthy that while having more empowerment is usually considered good for an agent, here there is a slight anticorrelation, indicating, if anything, that higher rated settlements have lower average reachability or empowerment. In part, this could be due to the fact that Minecraft settlements also aim to provide security, which they do by for example building houses that limit the way all agents, including monsters, can move around. If we look at the heatmap for reachability, we see that the typical houses in entry x reduce the reachability around them. Noteworthy is that the megastructure from x actually increases reachability, because it provides several levels of building, connected by stairs, allowing an agent to reach more locations than would be possible on flat land.

Occlusivity measure how much of the locations reachable in n steps are visible from the current position. Basically a measure of how easy it is to turn a corner and get lost. It is one of the measures that captures natural features as well - as forested areas and ragged mountains have more occlusivity (indicated by a lower value) then the open dessert in the north. Similarly, and expected, the cramped dystopian skyscraper with its complicated inner structure has a very high degree of occlusivity.

\section{Conclusion}

Looking back at our objective when we started this research, we were looking for generic automated metrics that would capture human perception of environmental features. This study is a first attempt at applying a well established theoretical tool, isovists, to PCG and video games. We introduced metrics based on isovistic properties, keeping them related to their primary purpose while adapting them to the Minecraft context. We transposed the idea of Perimeter and Area to a specific domain, and all their derivated metrics. We also derived metrics from the isovist's theory, specialized for virtual environment: Reachability and Clutter. Finally, the block type count metric was incorporated to a vision based evaluation called Diversity. Several of those metrics either correlated with human value judgement, or allowed for other insights into the problem. An additional output of this experiment is a fully functional and optimized framework dedicated to isovist measurement in Minecraft, that we can reuse in the future. 

The metrics introduce here can be applied to a wider range of 3d environments, both based on blocks, but possibly also to continuous environments by using the raycast as a discretization tool. The metrics are also local, allowing for the evaluation of a game state, or game-play trajectory, and a comparison with a reported experience. Unfortunately, there is not data set that contains human judgment or experiences for a given game state yet. As a result, we used average values that required us to test as much isovists as possible, regardless of their intrinsic properties (undergound, in the middle of a street, in an open field, ...), and average them. But as the heatmaps demonstrated, we can imagine that a generator may test the indoor of generated building, or key location of a settlement being created. Being able to spot weak points and irregularity, or in contrary uniformity, in content could be a way to improve a generator. Local metrics are a step further in understanding generated artifact in their entirety, rather than focussing on averaged value that try to capture the overall qualities of an asset - which might not be possible for complex artifacts in principle. 

Finally, we want to note again that the base sets used for the computation consider various embodiment properties of the player avatar in their computation - making them not only agent focused, but also opening interesting areas for further investigations towards their effects. By modeling an agents embodiment we try to make another step towards evaluating an agent's experience with and artifact, rather than the artifact itself. This might be a way out regarding the problem that none of the artifact based metrics are currently suitable for optimization - as having maximal Area, Perimeter and Block Count Type would still lead to a degenerative solution. On the other hand, optimizing for a given experience might prove more fruitful.

\bibliographystyle{ACM-Reference-Format}
\bibliography{sample-base}

\end{document}